\documentclass[letterpaper, 10 pt, conference]{ieeeconf}

\IEEEoverridecommandlockouts

\usepackage{color}
\usepackage{multirow}
\usepackage{booktabs}
\usepackage{url}
\usepackage{threeparttable}
\usepackage{cite}
\usepackage[ruled,linesnumbered]{algorithm2e}
\makeatletter
\let\NAT@parse\undefined
\makeatother
\usepackage{hyperref}[colorlinks,linkcolor=blue]
\renewcommand{\baselinestretch}{0.99}
\usepackage{siunitx}

\usepackage{enumerate}
\usepackage{amssymb}
\usepackage{leftidx}
\usepackage{amsfonts}
\usepackage{amsmath}
\usepackage{bm}
\usepackage{booktabs}
\usepackage{setspace}
\usepackage{makecell}
\usepackage{setspace}
\usepackage{float}
\usepackage[pdftex]{graphicx}
\graphicspath{{figs/}}
\usepackage{graphicx}
\graphicspath{{figs/}}
\newcommand{\etal}{\textit{et al}.}
\newcommand{\ie}{\textit{i}.\textit{e}.}
\newcommand{\eg}{\textit{e}.\textit{g}.}
\title{\LARGE \bf Swarm-LIO: Decentralized Swarm LiDAR-inertial Odometry}
\author{Fangcheng Zhu$^{*}$, Yunfan Ren$^{*}$, Fanze Kong, Huajie Wu, Siqi Liang, Nan Chen, Wei Xu, Fu Zhang
\thanks{$^*$These two authors contribute equally to this work.}
\thanks{F. Zhu, Y. Ren, F. Kong, H. Wu and F. Zhang are with Department of Mechanical Engineering, The University of Hong Kong. \{\textit{zhufc,renyf,kongfz,wu2020,cnchen}\}\textit{@connect.hku.hk}, \{\textit{xuweii,fuzhang}\} \textit{@hku.hk}. S. Liang is with Harbin Institute of Technology, Shenzhen \textit{sqliang@stu.hit.edu.cn}}
}

\usepackage{tikz}
\usetikzlibrary{shapes,snakes}

\begin{document}
\thispagestyle{empty} 
\pagestyle{empty}  
\maketitle
\begin{tikzpicture}[overlay, remember picture]
  \path (current page.north) ++(0.0,-1.0) node[draw = black] {Accepted by the 2023 IEEE International Conference on Robotics and Automation (ICRA), London};
\end{tikzpicture}
\vspace{-0.3cm}

\pagestyle{empty}  
\thispagestyle{empty} 
\begin{abstract}
Accurate self and relative state estimation are the critical preconditions for completing swarm tasks, \eg, collaborative autonomous exploration, target tracking, search and rescue. This paper proposes Swarm-LIO: a fully decentralized state estimation method for aerial swarm systems, in which each drone performs precise ego-state estimation, exchanges ego-state and mutual observation information by wireless communication, and estimates relative state with respect to (w.r.t.) the rest of UAVs, all in real-time and only based on LiDAR-inertial measurements. A novel 3D LiDAR-based drone detection, identification and tracking method is proposed to obtain observations of teammate drones. The mutual observation measurements are then tightly-coupled with IMU and LiDAR measurements to perform real-time and accurate estimation of ego-state and relative state jointly. Extensive real-world experiments show the broad adaptability to complicated scenarios, including GPS-denied scenes, degenerate scenes for camera (dark night) or LiDAR (facing a single wall). Compared with ground-truth provided by motion capture system, the result shows the centimeter-level localization accuracy which outperforms other state-of-the-art LiDAR-inertial odometry for single UAV system.
\end{abstract}

\section{Introduction}
Multi-robot system, especially aerial swarm system, has great potential in many aspects, such as autonomous exploration\cite{gao2022meeting, zhou2021fuel}, target tracking\cite{zhu2020multi}, search and rescue, etc. Thanks to its great team cooperation capability, a swarm system can complete missions in complex scenarios even in degenerate environment for a single drone.
For a single drone system, accurate self-localization\cite{xu2022fast,qin2018vins,lin2022r,r3live_pp} is the precondition for obstacle avoidance\cite{ren2022bubble,kong2021avoiding,ren2022online} and flight control\cite{lu2022model}. For an aerial swarm, robust, accurate self and relative state estimation\cite{xu2020decentralized} play a similarly vital role. For a swarm to fulfill cooperative tasks, each drone in the swarm needs to perform real-time, precise self-localization with low drift, and needs to be aware of other drones' states at all times.

Most recent works on aerial swarm mainly focus on collaborative planning\cite{zhou2022swarm,zhou2021ego,honig2018trajectory}, while research on self and relative state estimation of swarm system is still a large gap.
For outdoor scenes, GPS and RTK-GPS are usually adopted \cite{jaimes2008approach, moon2016outdoor}. While for GPS-denied places such as indoor scenarios, motion capture system\cite{honig2018trajectory}, visual (-inertial) odometry\cite{zhou2021ego, lajoie2020door} or LiDAR (-inertial) odometry\cite{gao2022meeting,dube2017online} methods are much preferred. Besides, Ultra-WideBand (UWB) is also adopted in \cite{zhou2022swarm,xu2022omni} to produce robust localization result.
Although RTK-GPS, motion capture and UWB with anchors \cite{zhou2021online} have great accuracy, they are dependent on cumbersome, extra devices and would lead to laborious installation. Moreover, these mentioned methods render the whole system centralized, which is fragile to single-point-of-failure (SPOF). 
Camera is widely used due to its light weight, low cost and rich color information, but it is vulnerable to inadequate illumination and is lack of direct depth measurement,
leading to high computational complexity when computing 3D measurements. Apart from that, the range of depth measurements of camera is quite limited.
Anchor-free UWB is also low-cost and light-weight. However, it can only offer one dimensional distance measurement and the accuracy is quite limited (usually meter-level), which may decrease the overall accuracy of the whole swarm system.

\begin{figure}[t]
    \setlength{\abovecaptionskip}{-0.1cm}
	\centering 
	\includegraphics[width=0.48\textwidth]{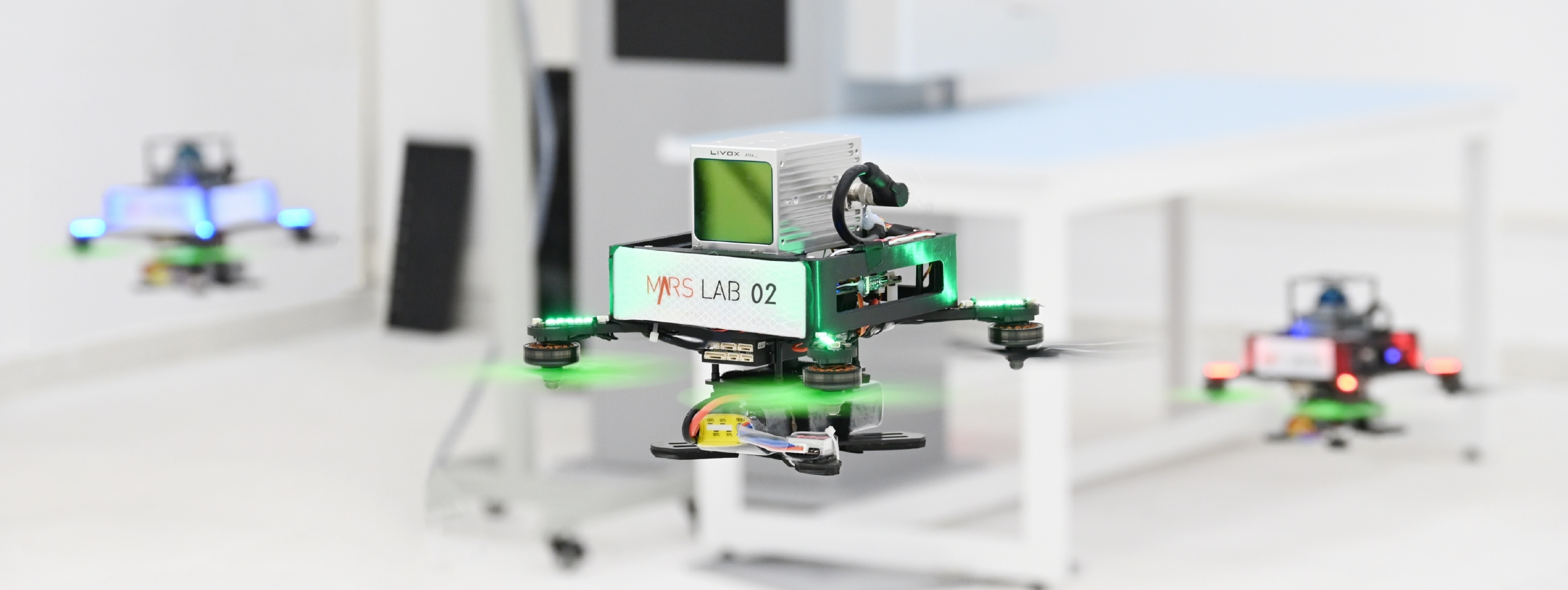}
	\caption{Indoor flights of the proposed aerial swarm with three drones.}
	\label{fig:cover}
\end{figure}

Compared with camera and anchor-free UWB, LiDAR can provide accurate 3D measurements and is also robust to illumination changes. Even in dark scenes, \eg~underground tunnels, LiDAR can help to perform accurate localization and mapping\cite{lajoie2020door}. Recent years, some cost-effective solid-state LiDARs such as Livox$\footnote{https://www.livoxtech.com}$ emerged in the market, which significantly expanded the scope of LiDAR-based researches.
In this paper, we propose a robust, real-time and decentralized swarm odometry based on LiDAR-inertial measurements. The main contributions are highlighted as follows:
\begin{itemize}
  \item [1)] A novel 3D LiDAR-based drone detection, identification and tracking method, providing accurate 3D mutual observation measurements used for self and relative state estimation. Each drone is attached with reflective tapes, such that a teammate drone can be reliably and efficiently detected from the reflectivity values of LiDAR points. The detected teammate drone is then tracked by a Kalman filter and matched with trajectories received from the shared network to obtain the teammate's identity and its initial relative state. 
  \item [2)] 
  A fully decentralized, robust, data-effective and computationally-efficient ego-state and relative state estimation framework. In this framework, each drone only has to exchange its ego-state and teammate observation information, which requires extremely low communication bandwidth. The mutual observation measurement is tightly-coupled with LiDAR and IMU measurements under Error State Iterated Kalman Filter (ESIKF) \cite{he2021kalman} framework to achieve robust, accurate, and simultaneous self and relative state estimation.
  \item [3)]
  Low-cost, decentralized hardware and software framework. The sensors and computing resource are totally on-board, including Livox LiDAR, IMU and on-board computer. The communication is achieved with Ad-Hoc network available on standard network module.
  \item [4)] Implementation and validation of the proposed method in extensive real-world experiments, in both indoor (see Fig.~\ref{fig:cover}), outdoor and degenerate scenes.
\end{itemize}


\section{Related Works}
State estimation for aerial swarm systems is much more complicated than that of single drone system because each drone not only needs to perform accurate self-localization with low drift, but also needs to estimate teammates' states in its own global frame. Among these swarm systems, cameras and LiDARs are two mainstream sensors.

Many existing works on swarm state estimation are based on image measurements.
In \cite{zhou2021ego,weinstein2018visual} each drone in the swarm runs an independent VIO to achieve ego-state estimation. The ego-state is then exchanged so that each drone can estimate the relative state of other teammates given the relative transformation between their initial pose (global extrinsic). The global extrinsic is usually calibrated offline\cite{zhou2021ego} or estimated online with good initial guess\cite{xu2020decentralized}. Since the VIO of each drone is completely independent, it may lead to large relative state estimation error as the system runs, due to imprecise global extrinsic or VIO drift caused by long-time running. To constrain the relative state during the online running of swarm systems, mutual observation based methods are vigorously developing.
Nguyen \etal \cite{nguyen2020vision} propose visual-inertial multi-drone localization system, MAVNet is used to detect other teammate drones. 
To further enhance the mutual observation, Xu \etal \cite{xu2020decentralized} propose a visual-inertial-UWB relative state estimation system utilizing YOLOv3-tiny for teammate drone detection, and UWB module is adopted to provide distance constraints as an complementary sensor of camera.
These learning-based drone detection methods usually needs long time of prior training, resulting in increasement of the total time-consumption of system implementation.
One of the biggest restrictions of vision-based methods is the range and quality of depth measurement are quite limited, which narrows down their applications in practical cases, \eg~in large-scale, outdoor environments.

By contrast, LiDAR can provide accurate and long-range depth measurements, bringing many new opportunities for swarm state estimation. In \cite{dube2017online, zhou2021online, denniston2022loop}, 3D LiDAR-based place recognition (loop closure) constraints are utilized to improve state estimation accuracy. However, since cloud register of raw or segmented points are inevitable in place recognition, the computation load is quite heavy. Thus, all \cite{dube2017online, zhou2021online, denniston2022loop} need a centralized master agent to receive data transmitted by other agents and finish those computationally intensive tasks, \ie, cloud register or map merging, increasing the risk of single-point-of-failure.
To ease the burden of communication and computation, Huang \etal \cite{huang2021disco} propose a decentralized multi-robot system using Scan Context descriptor, permitting data-efficient exchange of LiDAR descriptors among robots. Apart from map and descriptor exchange mentioned above, mutual observation is another important unique feature of swarm system, which is a light-weight type of information, avoiding large communication and computation load.
Wasik \etal \cite{wasik2016lidar} propose a laser-based multi-robot system, laser range finders are used for each robot to estimate the distances and angles to other robots. However, the used 2D LiDARs are not applicable to drones considered in this paper which fly in 3D spaces. 

Compared with learning-based detection methods \cite{xu2020decentralized,nguyen2020vision}, we attach reflective tapes on each drone and leverage the reflectivity measured by LiDAR sensors to detect teammate drones. Different from \cite{xu2020decentralized, zhou2021ego}, we propose a course-to-fine calibration method to acquire accurate global extrinsic transformations without any prior initial guess. The rough calibration result acquired by trajectory matching is fed into ESIKF for further online refinement along with ego-state estimation. Compared with centralized systems in which the drones exchange map information \cite{dube2017online, zhou2021online, denniston2022loop}, our system is fully decentralized which would suffer no SPOF issue, and communication-efficient which exchanges ego-state and mutual observation information only.

\begin{figure*}[t]
    \setlength{\abovecaptionskip}{-0.1cm}
	\centering 
	\includegraphics[width=0.85\textwidth]{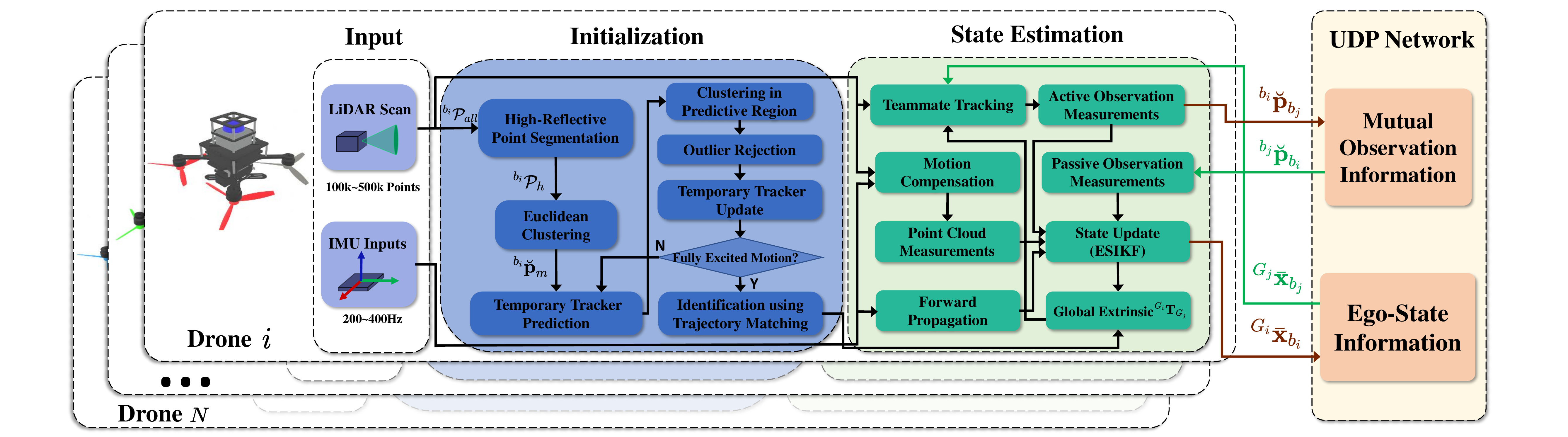}
	\caption{Framework of the proposed state estimation system for aerial swarm systems.}
	\label{fig:framework}
\end{figure*}

\section{Methodology}
\subsection{Problem Formulation}\label{section:problem_formulation}
Aiming to assist understanding of the proposed system, we define some important notations here.
$\mathbf x_i,\widehat {\mathbf x}_i, \bar{\mathbf x}_i, \breve{\mathbf x}_i$ represent the ground-truth, predictive, updated, and measured state of drone $i$ respectively. $t_{ik}$ denotes the timestamp of the $k$-th ESIKF state update of drone $i$.

Considering an aerial swarm system consisting of $ N$ drones and each one carries a LiDAR and a inertial measurement unit (IMU), the state estimation is decomposed into two parts. 
The first part is ego-state estimation. Each drone, here we choose drone $i$ for demonstration convenience, needs to estimate its own position $^{G_i}\mathbf p_{b_i} \in \mathbb R^{3}$ and attitude $^{G_i}\mathbf R_{b_i} \in SO(3)$ in its global frame $G_i$. Since the IMU measurements are coupled with unknown and time-varying bias, the gyroscope bias $\mathbf b_{g_i} \in \mathbb R^{3}$ and accelerometer bias $\mathbf b_{a_i}\in \mathbb R^{3}$ should be calibrated. Moreover, gravity vector in each global frame $^{G_i}\mathbf g \in \mathbb R^{3}$ should also be estimated online. The self position  $^{G_i}\mathbf p_{b_i}$, attitude $^{G_i}\mathbf R_{b_i}$, velocity  $^{G_i}\mathbf v_{b_i}$, bias $\mathbf b_{g_i}, \mathbf b_{a_i}$, and gravity vector $^{G_i}\mathbf g$ constitute the ego-state, whose estimation is exchanged among the swarm system. Then, to estimate the relative state of other teammate drones based on the exchanged ego-state information, each drone needs to estimate the global extrinsic transformation ${^{G_i}}\mathbf T_{G_j} = ({^{G_i}}\mathbf R_{G_j},{^{G_i}}\mathbf p_{G_j}) \in SE(3)$ with respect to the teammate drone $j$ $(j = 1,\cdots, N, j \neq i)$.
Finally, to provide mutual observations of other teammates, each drone $i$ needs to detect, identify, and track any other teammate $j$, whose observed position in the self body frame is denoted by ${^{b_i}}\breve{\mathbf p}_{b_j}$. 

\subsection{Framework Overview}
Each drone in the swarm system runs an independent copy of the same framework.  The overview of the framework copy run by the $i$-th drone is shown Fig.~\ref{fig:framework}. The first stage is initialization (Sec. \ref{section:initialization}), including detection and temporary tracking for potential teammate drones (Sec. \ref{section:detection}). Once the motion of temporarily tracked object is sufficiently excited, the trajectory matching (Sec. \ref{section:traj_match}) starts. If the trajectory of the temporarily tracked object is successfully matched with a teammate's trajectory transmitted by wireless communication, the object would be considered as the observation of the corresponding teammate and be labeled with the teammate identity (ID), the temporary tracker hence becomes a teammate tracker (Sec. \ref{section:teammate_tracking}). The observed position of teammate $j$, ${^{b_i}}\breve{\mathbf p}_{b_j}$ (the ``active observation measurement"), the self-position observed by teammate $j$, ${^{b_j}}\breve{\mathbf p}_{b_i}$ ( the ``passive observation measurement", which is received from the network), the LiDAR point cloud (after motion compensation), and the IMU measurements are then fused  by an error-state iterative Kalman filter (ESIKF) (Sec. \ref{section:state estimation}) to jointly estimate the ego-state and global extrinsic transformation ${^{G_i}}\mathbf T_{G_j}$.
Finally, the estimated ego-state and active mutual observation information are transmitted to other drones via UDP packets by Ad-Hoc communication.

\subsection{Initialization of Swarm System} \label{section:initialization}
In this section, we introduce how the drones detect, track and identify a potential teammate drone.
\subsubsection{Drone Detection and Temporary Tracking} \label{section:detection}
 We propose a novel drone detection method based on reflectivity filtering and cluster extraction. For each drone, reflective tapes are attached to its body, so that it can be easily detected by other teammates based on the reflectivity information measured by LiDAR sensor. The detailed detection and tracking procedures at each LiDAR scan for drone $i$ are summarized in Alg.~\ref{alg:initialization}. Denote ${^{b_i}} \mathcal{P}_{all}$ as the raw LiDAR points in a new scan represented in the current body frame. First, points with high reflectivity values exceeding a given threshold, which can be calibrated beforehand on the reflective tapes, are extracted by \textbf{ReflectivityFiltering}(${^{b_i}} \mathcal{P}_{all}$) in Line \ref{alg:ReflectivityFiltering}. Then in Line \ref{alg:EuclideanClustering1}, the high-reflectivity points ${^{b_i}} \mathcal P_h$ are clustered by \textbf{EuclideanClustering}(${^{b_i}} \mathcal P_h$)$\footnote{https://pcl.readthedocs.io/en/latest/cluster\_extraction.html}$ to detect new potential teammate drones. 
 The newly detected object is then tracked by a Kalman filter-based temporary tracker in Line \ref{alg:TemporaryTracking}.

  The state vector $\mathbf x_m$ of each temporary tracker $m$ ($m=1,2,\cdots M$) consists of the object position ${^{G_i}}\mathbf p_{m}$ and velocity in drone $i$'s global frame and the temporary trackers will predict a tracker position ${^{G_i}}\hat{\mathbf p}_{m}$ based on constant velocity model. Then, the position ${^{G_i}}\breve{\mathbf p}_{m} \!=\! {^{G_i}}\mathbf T_{b_i} {^{b_i}}\breve{\mathbf p}_m$ (where ${^{b_i}}\breve{\mathbf p}_m$ is clustered from high-reflectivity points in Line \ref{alg:EuclideanClustering1}) is associated with the closest predicted position. If no valid association is found, \ie, error of predicted position and cluster position is too large, the tracker re-clusters the raw points around the predicted position ${^{G_i}}\hat{\mathbf p}_{m}$. The reason is that the measured point reflectivity values are often affected by the object distance and laser incidence angle, so the extracted high-reflectivity points may not represents all points on the potential teammate drone. To obtain a more accurate cluster, we calculate a {\it predicted region } centered at ${^{G_i}}\hat{\mathbf p}_{m}$ and cluster an object within this region, which is then used to update the temporary tracker. Noted that the extracted objects whose size is much larger or smaller than actual drone size are rejected (invalid object).
  If no valid object can be clustered, the tracker will propagate to the next step. The tracker will be killed if it has propagated for too many steps without an update. Note that since the points in predicted region are far less than all input points, the time consumption of re-clustering would be significantly decreased.

 \begin{algorithm}[t]
  \footnotesize
  \caption{Initialization}
  \label{alg:initialization}
  \KwIn{
    Raw LiDAR points in body frame ${^{b_i}} \mathcal P_{all}$,
    drone $i$'s odometry ${^{G_i}}\mathbf T_{b_i}$, time interval of LiDAR input $\Delta t$, drone $j$'s position trajectory $^{G_j}\mathcal T_j$, given threshold of trajectory matching residual $thr$.
  }
  \KwOut{
    Global extrinsic transformation ${^{G_i}}\mathbf T_{G_j}$, tracked position of drone $j$ in drone $i$'s global frame ${^{G_i}} \mathbf p_{b_j}$
  }  
  \BlankLine
  ${^{b_i}} \mathcal P_h$ = ReflectivityFiltering(${^{b_i}} \mathcal P_{all}$)\; \label{alg:ReflectivityFiltering}
  ${^{b_i}} \breve{\mathbf p}_m$ = EuclideanClustering(${^{b_i}} \mathcal P_h$)\; \label{alg:EuclideanClustering1}
  \For{$m = 1:M$}{
    ${^{G_i}} \bar{\mathbf p}_m$ = TemporaryTracker($\Delta t, {^{G_i}}\mathbf T_{b_i}, {^{b_i}} \mathcal P_{all}, {^{b_i}} \breve{\mathbf p}_m$)\; \label{alg:TemporaryTracking}
    $^{G_i}\mathcal T_m$.PushBack(${^{G_i}} \bar{\mathbf p}_m$)\;

    \If(){TrajExcited($^{G_i}\mathcal T_m$) \label{alg:TrajExcited}}
    {
      \For{$ = 1:N$; $j \neq i$}
      {
        $res, \mathbf T$ = TrajMatching(${^{G_j}}\mathcal T_j, {^{G_i}}\mathcal T_m$)\; \label{alg:TrajMatching}
        \If(){$res < thr$}
        {
          ${^{G_i}}\mathbf T_{G_j} = \mathbf T$ \;
          ${^{G_i}} \mathbf p_{b_j}$ = ${^{G_i}} {\mathbf p}_m$\;
          break\;
        }
      }
    }
  }
\end{algorithm}

\subsubsection{Teammate Identification using Trajectory Matching} \label{section:traj_match}
By rejecting invalid clusters and tracking real potential teammates, the trajectory of each temporary tracker is accumulated for subsequent identification. Since all drones in the proposed swarm system would exchange their estimated ego-state (in their own global frame) with others, the teammate identification and global extrinsic can be acquired by trajectory matching as follows:
\begin{equation}\label{trajmatching}
    \arg\min_{^{G_i}\mathbf T_{G_j}}\sum_{\kappa = 1}^\mathcal K \dfrac{1}{2}\| {^{G_i}}\bar{\mathbf p}_{m,\kappa} - {^{G_i}}\mathbf T_{G_j}{^{G_j}} \breve{\mathbf p}_{b_j,\kappa}  \|
\end{equation}
where ${^{G_i}}\bar{\mathbf p}_{m,\kappa} \! \in \! {^{G_i}}{\mathcal T}_{m}$ represents the $\kappa$-th position in the tracked position trajectory ${^{G_i}}{\mathcal T}_{m}$ of the $m$-th object and ${^{G_j}} \breve{\mathbf p}_{b_j,\kappa} \! \in \! {^{G_j}}{\mathcal T}_{j}$ represents position received from drone $j$. Considering possible short-term communication disconnection, some data of ${^{G_j}} \breve{\mathbf p}_{b_j}$ might be lost. Thus, we only pick ${^{G_i}}\bar{\mathbf p}_{m,\kappa}$ that has close timestamp with ${^{G_j}} \breve{\mathbf p}_{b_j,\kappa}$ to participate in trajectory matching. Besides, to avoid large computing time due to too much data, we use a sliding window of the most recent $\mathcal K$ positions for matching.

Since no unique transformation can be determined from (\ref{trajmatching}) if the involved trajectories are straight lines\cite{PointRegister}, the trajectories of those tracked objects are constantly assessed by \textbf{TrajExcited}($^{G_i}\mathcal T_m$) in Line \ref{alg:TrajExcited} until sufficient information is collected. 
Let ${^{G_i}} \bar {\mathbf p}_{m}^c$ represent the centroid of $^{G_i}\mathcal T_m$, \textbf{TrajExcited}($^{G_i}\mathcal T_m$) assess the excitation (shape) of $^{G_i}\mathcal T_m$ by computing the singular values of matrix $\mathcal H \in \mathbb R^{3\times 3}$:
\begin{equation}
    \mathcal H \triangleq \sum_{\kappa=1}^\mathcal K ({^{G_i}} \bar{\mathbf p}_{m,\kappa} - {^{G_i}} \bar{\mathbf p}_{m}^c)\cdot ({{^{G_i}} \bar{\mathbf p}_{m,\kappa}} - {^{G_i}} \bar{\mathbf p}_{m}^c)^T
\end{equation}

\begin{figure}[t]
    \setlength{\abovecaptionskip}{-0.1cm}
	\centering 
	\includegraphics[width=0.45\textwidth]{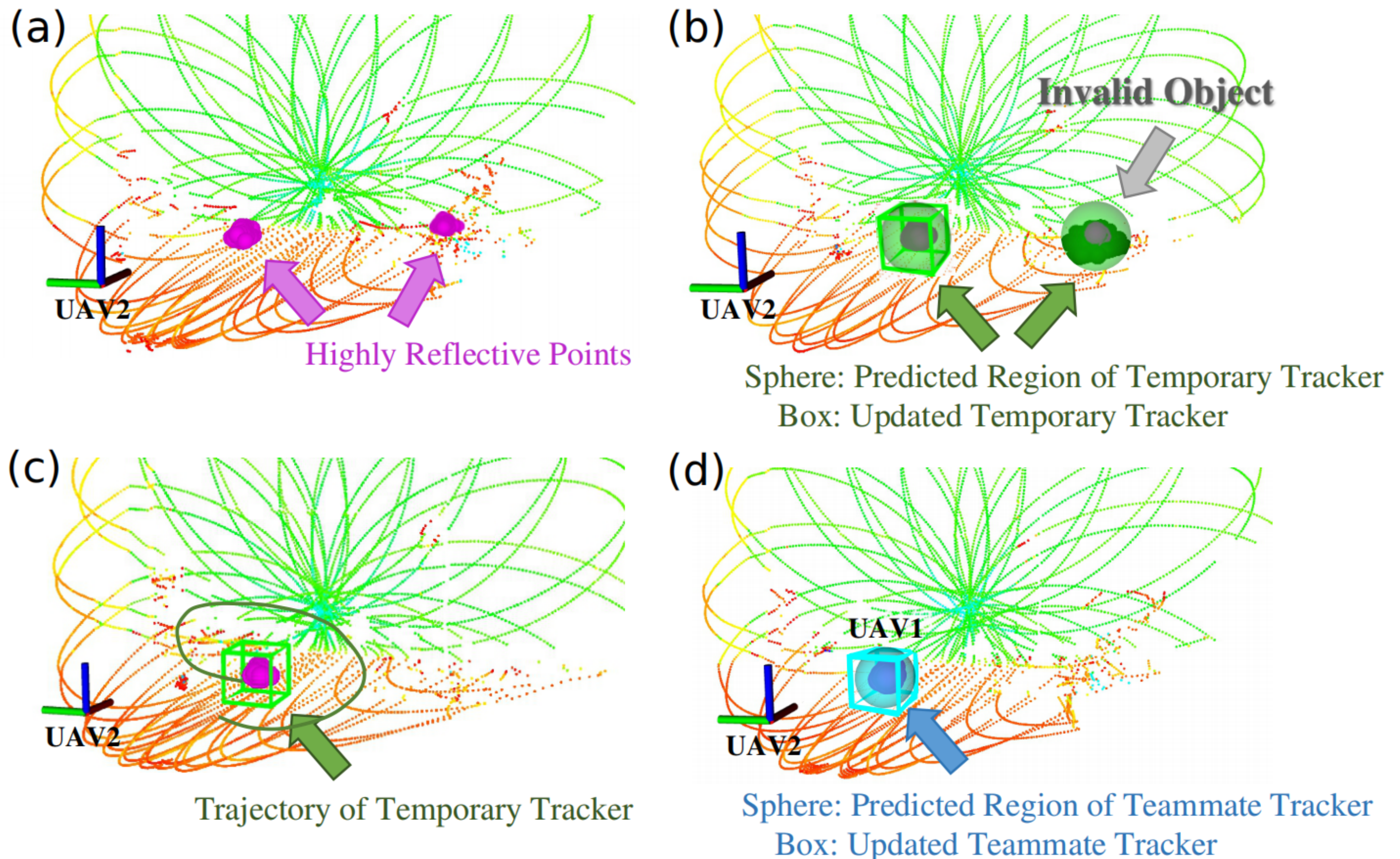}
	\caption{Illustration of the initialization for newly detected objects, point-cloud is colored by reflectivity. Here the self drone is UAV2 and it needs to detect and identify other teammate UAVs in its FoV. Center of the box represents the updated position of the tracker. (a) Reflectivity filtering. (b) Outlier rejection by discarding objects with too large size. (c) Track real potential teammates and accumulate the trajectory. (d) After trajectory matching, the object is identified as UAV1, and the temporary tracker becomes a teammate tracker (see Sec. \ref{section:teammate_tracking}).}
	\label{fig:initialization}
\end{figure}

If the second largest singular value is larger than a given threshold, the trajectory is fully excited, which is qualified to do \textbf{TrajMatching}(${^{G_j}}\mathcal T_j, {^{G_i}}\mathcal T_m$) in Line \ref{alg:TrajMatching}. This function solves (\ref{trajmatching}) which has well-studied closed-form solution \cite{PointRegister}. The matching is performed with each received teammate drone's trajectory until the matching error is smaller than a given threshold, indicating that the object $m$ is essentially the observation of teammate $j$, and the solution of \eqref{trajmatching} gives an initial estimation of the global extrinsic ${^{G_i}}\mathbf T_{G_j}$, which is then refined online using the ESIKF in Sec. \ref{section:esikf}. After identification, a temporary tracker becomes teammate tracker with the corresponding drone ID, which will be sequentially tracked as Sec. \ref{section:teammate_tracking}. The initialization pipeline is illustrated in Fig.~\ref{fig:initialization}.

\subsection{Decentralized LiDAR-inertial State Estimation}\label{section:state estimation}
The fully decentralized state estimation of the proposed swarm system is a tightly-couple iterated Kalman filter inherited from FAST-LIO2 \cite{xu2022fast}, but further incorporates the mutual observation constraints to improve ego-state estimation accuracy, and includes online refinement of inter-drone global extrinsic transformations. 

\subsubsection{Teammate Tracking} \label{section:teammate_tracking}
After the detection and identification of a teammate drone, a teammate tracker is obtained and the global extrinsic is initially calibrated by trajectory matching. The teammate tracker is similar to a temporary tracker, but with two key differences.
The first difference lies in the prediction. Both temporary and teammate tracker predict state using constant velocity model, but a teammate tracker predicts the state based on the velocity received from the corresponding teammate (and the most recent estimation of the corresponding global extrinsic transformation), instead of from a ego-estimated velocity in a temporary tracker (see Sec. \ref{section:detection}).
The second difference lies in state update when no teammate observation is available. In a temporary tracker, if there is no valid observation, \eg, the teammate is outside FoV, the tracker will propagate for a few steps and then terminate as explained in Sec. \ref{section:detection}. In a teammate tracker, it will use the teammate odometry received from the network, after transforming to the self drone's global frame using the most recent global extrinsic transformation obtained by Sec. \ref{section:esikf}, to continue the state update. 

\subsubsection{State Prediction} 
Denote $\tau$ as the IMU measurement index, the discrete state transition model is shown below:
\begin{equation}\label{discrete_model}
    \mathbf x_{i,\tau+1} = \mathbf x_{i,\tau} \boxplus(\Delta t_\tau \mathbf f_i(\mathbf x_{i,\tau}, \mathbf u_{i,\tau}, \mathbf w_{i,\tau}))
\end{equation}
 $\Delta t_\tau$ is the time interval of two consecutive IMU measurements, $\mathbf x_{i,\tau}$ denotes ground-truth of state at IMU measurement's timestamp $t_{i\tau}$. The state vector $\mathbf x_i$, discrete state transition function $\mathbf f_i$, noise $\mathbf w_i$ and input $\mathbf u_i$ are defined as:
\begin{equation*}\label{kinematic_model}
\begin{aligned}
\mathbf x_i &\triangleq
[
^{G_i}\mathbf R_{b_i} \ 
^{G_i}\mathbf p_{b_i}\ 
^{G_i}\mathbf v_{b_i}\ 
\mathbf b_{g_i}\ 
\mathbf b_{a_i}\ 
^{G_i}\mathbf g \\
& \qquad \cdots \quad
^{G_i}\mathbf R_{G_j}\ 
^{G_i}\mathbf p_{G_j}\quad
\cdots 
]\in \mathcal M\\
\mathbf f_i  &\triangleq
[
\omega_{m_i}  \!-\! \mathbf b_{g_i}  \!-\!\mathbf n_{g_i}\
^{G_i}\mathbf v_{b_i} \ 
^{G_i}\mathbf R_{b_i} (\mathbf a_{m_i} \!-\!\mathbf b_{a_i} \!- \!\mathbf n_{a_i}) \!+\! {^{G_i}}\mathbf g\\
& \qquad
\mathbf  n_ {\mathbf b_{g i}}\ 
\mathbf n_{\mathbf b_{\mathbf a i}}\ 
\mathbf 0_{3\times 1} \ 
\cdots \
\mathbf n_{R}\ 
\mathbf n_{p}\
\cdots
]\\
\mathbf w_i  \! &\triangleq \! 
\begin{bmatrix}
\mathbf n_{g_i} \! \!&\! \!
\mathbf n_{a_i}  \! \! & \! \!
\mathbf  n_ {\mathbf b_{gi}} \! \!& \! \!
\mathbf n_{\mathbf b_{ai}}   \! \!& \! \!
\mathbf n_{R}   \! \!& \! \!
\mathbf n_{p} 
\end{bmatrix},\quad
\mathbf u_i \! \triangleq \! 
\begin{bmatrix} 
\omega_{m_i}  \! \!& \! \!
\mathbf a_{m_i} 
\end{bmatrix}
\end{aligned}
\end{equation*}
where $\omega_{m_i}, \mathbf a_{m_i}$ represent the IMU measurements of drone $i$, the meaning of each element in state vector $\mathbf x_i$ is introduced in Sec. \ref{section:problem_formulation}, the state manifold $\mathcal M$ are defined in \eqref{dim_M} and its dimension is $18+6\times( N-1)$.
\begin{equation}\label{dim_M}
    \mathcal M \triangleq \underbrace{SO(3) \times \mathbb R^{15}}_{\text{dim} = 18} \times \underbrace{\cdots \times  SO(3) \times \mathbb R^{3} \times \cdots}_{\text{dim} = 6 \times ( N-1)}
\end{equation}

In (\ref{discrete_model}), we used the notation $\boxplus/\boxminus$ defined in \cite{hertzberg2013integrating} to compactly represent the ``plus" on the state manifold. Specifically, for the state manifold $SO(3) \times \mathbb R^n$ in (\ref{kinematic_model}), the $\boxplus$ operation and its inverse $\boxminus$ are defined as
\begin{equation*}
    \begin{bmatrix}
    \mathbf R \\ \mathbf a
    \end{bmatrix} \boxplus
    \begin{bmatrix}
    \mathbf r \\ \mathbf b
    \end{bmatrix}=
    \begin{bmatrix}
    \mathbf R\text{Exp}(\mathbf r) \\ \mathbf {a+b}
    \end{bmatrix};
    \begin{bmatrix}
    \mathbf R_1 \\ \mathbf a
    \end{bmatrix} \boxminus
    \begin{bmatrix}
    \mathbf R_2 \\ \mathbf b
    \end{bmatrix}=
    \begin{bmatrix}
    \text{Log}(\mathbf R_2^T \mathbf R_1) \\ \mathbf {a-b}
    \end{bmatrix}
\end{equation*}
where $\mathbf R, \mathbf R_1, \mathbf R_2 \in SO(3),\mathbf r \in \mathbb R^3 , \mathbf {a,b} \in \mathbb R^n$, $\text{Exp}(\cdot): \mathbb{R}^3 \mapsto SO(3) $ is the exponential map on $SO(3)$\cite{hertzberg2013integrating} and $\text{Log}(\cdot): SO(3) \mapsto \mathbb{R}^3$ is its inverse logarithmic map.

The state prediction step of the $i$-th drone's state under ESIKF framework is implemented once receiving a new IMU measurement as follows:
\begin{equation}\label{eq:state_prediction}
    \widehat {\mathbf x}_{i,\tau+1} = \widehat {\mathbf x}_{i,\tau} \boxplus(\Delta t_\tau \mathbf f_i(\widehat {\mathbf x}_{i,\tau}, \mathbf u_{i,\tau}, \mathbf 0));\widehat {\mathbf x}_0 = \bar {\mathbf x}_{i,k-1}
\end{equation}

\subsubsection{Error State Iterative State Update}\label{section:esikf}
The update step is implemented iteratively at the end time of new LiDAR scan $t_{ik}$, fusing point-cloud measurements and mutual observation measurements (if any). Once receiving a new scan, motion compensation would be performed to obtain undistorted points, and point-to-plane distance will be computed to generate point-cloud residuals. Details of the motion compensation can be referred to \cite{xu2022fast}. Denote each motion undistorted point projected into global frame using propagated ego-state as $^{G_i}\widehat {\mathbf p}_{n}$ , denote $\mathbf u_n$ as the normal vector of the corresponding plane, on which lying a point ${^{G_i}}\mathbf q_{n}$, the point residual is denoted as $\mathbf z_{p,n}$.

Apart from point-cloud residual, one main contribution of this paper is 3D LiDAR-based mutual observation measurements, which are used to construct new constraints to improve state estimation accuracy and to render the swarm system robust to degenerate scenes. For drone $i$, denote $\mathbf z_{ao,ij}$ as the active observation residual arising from ${^{b_i}}\breve{\mathbf p}_{b_j}$ (the active observation measurement w.r.t drone $j$, see Sec. \ref{section:teammate_tracking}), and denote $\mathbf z_{po,ij}$ as the passive observation residual arising from ${^{b_j}}\breve{\mathbf p}_{b_i}$ (the passive observation measurement w.r.t drone $j$), then the residual block of drone $i$ is composed as:
\begin{equation}\label{residual_block}
\begin{aligned}
    \mathbf z_i &= [\cdots,\mathbf z_{p,n}^T,\cdots, \mathbf z_{ao,ij}^T,\cdots, \mathbf z_{po,ij}^T,\cdots]^T \\
    \mathbf z_{p,n} &= \mathbf u_n^T( {^{G_i}}\widehat {\mathbf p}_{n} - {^{G_i}}\mathbf q_{n})\\
    \mathbf z_{ao,ij} &= {^{G_i}}\mathbf T_{b_i}^{-1} {^{G_i}}\mathbf T_{G_j}\!{^{G_j}}\breve{\mathbf p}_{b_j} - {^{b_i}}\breve{\mathbf p}_{b_j}\\
    \mathbf z_{po,ij} &= {^{G_j}}\breve{\mathbf T}_{b_j}^{-1} {^{G_i}}\mathbf T_{G_j}^{-1}{^{G_i}}\mathbf p_{b_i} - {^{b_j}}\breve{\mathbf p}_{b_i} \\
\end{aligned}
\end{equation}

The state will be iteratively updated until convergence, the details can be referred to \cite{xu2022fast}. After convergence, the map will be incrementally updated using ikd-tree \cite{cai2021ikd}, an efficient data structure for map management. The relative state estimation is completed by projecting the ego-state information transmitted by teammate drones using the updated global extrinsic transformations.

\section{Experiments}\label{experiments}
\begin{figure}[h]
    \setlength{\abovecaptionskip}{-0.1cm}
	\centering 
	\includegraphics[width=0.42\textwidth]{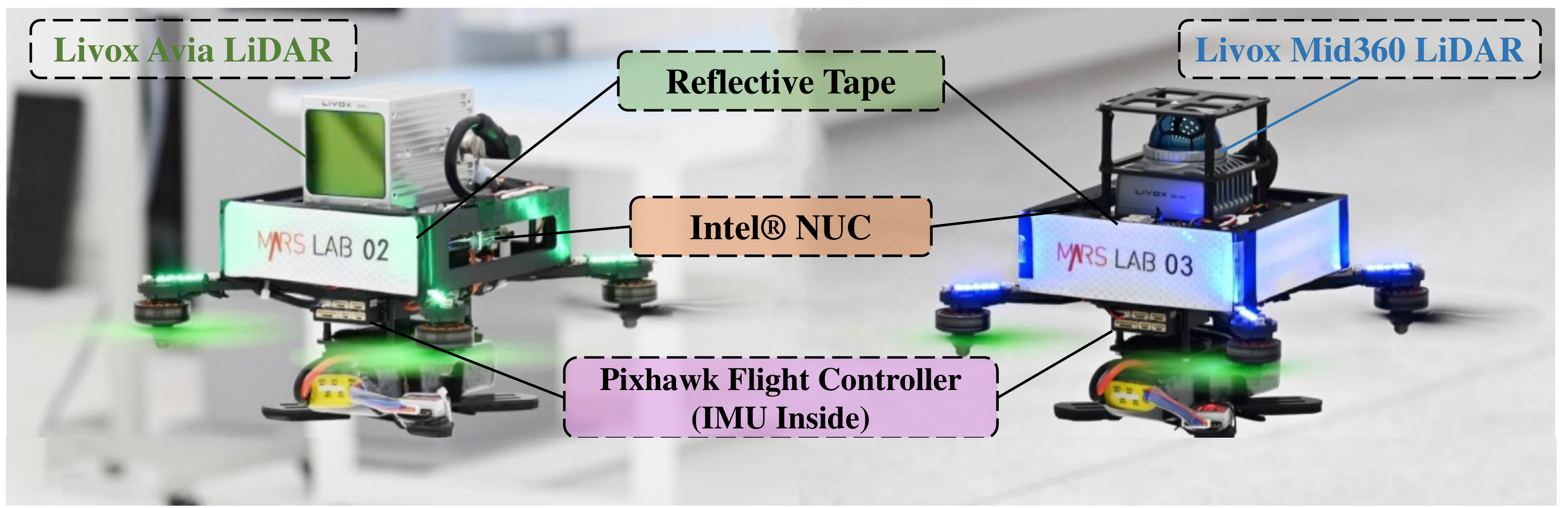}
	\caption{The UAV platform of the proposed swarm system. Since the device setup of drone 1 is the same as drone 3, only the picture of drone 3 is shown.}
	\label{fig:platform}
\end{figure}

The experiment platform contain three drones, and each one carries a Livox LiDAR (drone 1 and drone 3 carry Livox Mid360 and drone 2 carries Livox Avia with a smaller FoV), a pixhawk flight controller (BMI055 6-axis IMU inside), an onboard computer Intel NUC with CPU i7-10710U. Each drone is attached with reflective tapes for easy detection (See Fig.~\ref{fig:platform}). For each drone, the time offset and extrinsic between LiDAR and IMU are calibrated by \cite{zhu2022robust}, and the time among different drones are synced beforehand based on Network Time Protocol (NTP). All drones fly in manual mode using model predictive control \cite{lu2022model}.

\subsection{Indoor Flights}
Five flights in indoor scenes are performed to evaluate the localization accuracy. Motion capture system is introduced as the ground-truth. As far as we know, since there is no other 3D LiDAR-based state estimation methods for aerial swarm systems, we conduct ablation study by comparing localization accuracy of our method to FAST-LIO2\cite{xu2022fast} (state-of-the-art LiDAR-inertial odometry for single drone system). From the position RMSE shown in Table~\ref{localization_accuracy}, it can be seen that for drone 2, the localization error of the proposed method fusing mutual observation measurements is obviously smaller than that of FAST-LIO2. The reason lies in the limited FoV of Livox Avia LiDAR, which is only 70.4$^\circ$ $\times$ 77.2$^\circ$. The small FoV makes the LiDAR easier to drift due to lack of structural features, especially in indoor scenes. 
In this case, the mutual observation measurements can provide strong constraints to acquire accurate localization results.
While for drone 1 and drone 3, thanks to the 360$^{\circ}$ horizontal FoV of Mid360 LiDAR, there are sufficient point-cloud constraints to keep FAST-LIO2 work well. Thus, fusing mutual observations does not improve the localization accuracy significantly.
For all experiments the time interval of LiDAR input is \SI{100}{\ms}, the average time-consumption per scan of the proposed method and FAST-LIO2 is \SI{17.5}{\ms} and \SI{14.8}{\ms}. It proved the proposed Swarm-LIO can be implemented in real-time and has approximate computational efficiency to FAST-LIO2 (a fast odometry for single UAV system).

\subsection{Degenerate Environment Flights}\label{section:degenerate}
\begin{figure}[t]
    \setlength{\abovecaptionskip}{-0.1cm}
	\centering 
	\includegraphics[width=0.45\textwidth]{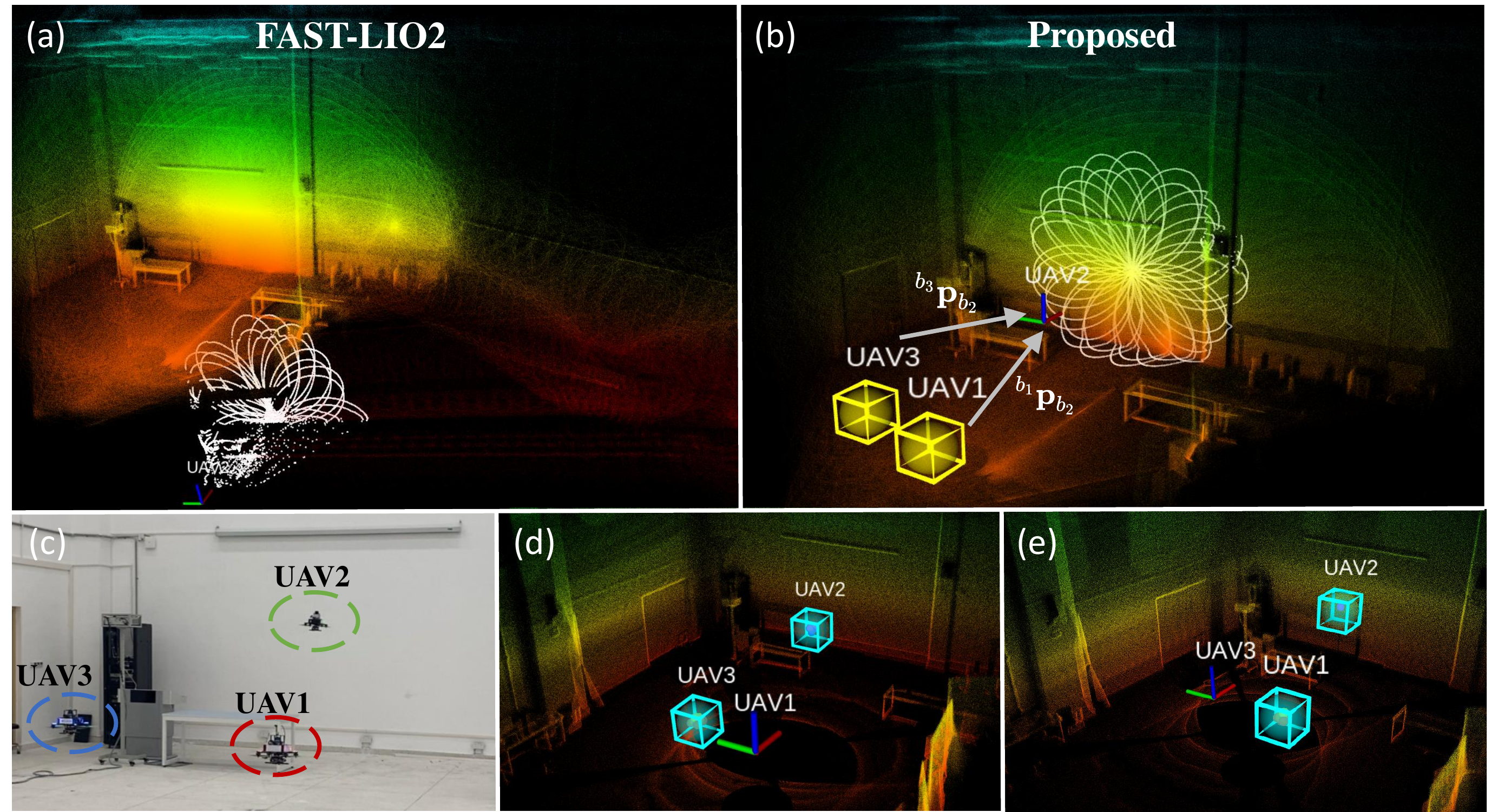}
	\caption{Degenerate smooth wall experiment. The white points in (a) and (b) refer to current input points. (a) View of drone 2 with point-cloud map constructed by FAST-LIO2, severe drift occurs and the map is messy. (b) View of drone 2 with point-cloud map constructed by the proposed method fusing passive observation measurements ${^{b_1}}\mathbf p_{b_2}$ and ${^{b_3}}\mathbf p_{b_2}$. Yellow boxes means that teammate drones are out of FoV, teammate trackers are updated fusing ego-state estimation transmitted by teammates. (c) Picture of the three drones. (d)(e) View of drone 1 and drone 3 respectively and their observations of teammate drones (the blue boxes).}
	\label{fig:degeneration1}
\end{figure}

The first degenerate scene is data 05 shown in Table~\ref{localization_accuracy}, where drone 2 carrying a Livox Avia LiDAR faces a smooth wall which imposes insufficient constraints to determine the full state. As a result, the single-agent odometry FAST-LIO2 completely degenerates with large RMSE. While for the proposed swarm system, drone 2 is observed by its teammate drone 1 and drone 3 (the passive observation measurement). Fusing passive observation measurements transmitted from drone 1 and drone 3 in \eqref{residual_block} will still lead to stable and accurate state estimation. The point-cloud map, mutual observation measurements are illustrated in Fig.~\ref{fig:degeneration1}.

\begin{figure}[t] 
    \setlength{\abovecaptionskip}{-0.1cm}
	\centering 
	\includegraphics[width=0.45\textwidth]{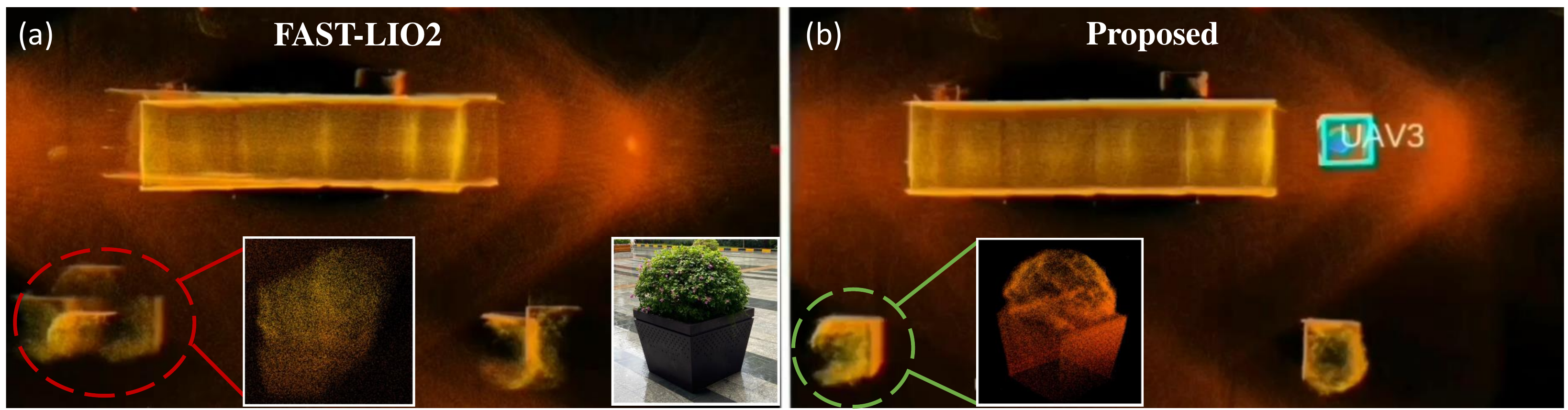}
	\caption{Point-cloud map of smooth cuboid corridor scene constructed by FAST-LIO2 (a) and the proposed method (b).}
	\label{fig:degeneration2}
\end{figure}

The second degenerate scene is a smooth cuboid corridor. Similar to the smooth wall, observations of a single LiDAR do not provide sufficient constraints for pose determination, so FAST-LIO2 suffers from drifts. While for a swarm system, when the first drone flies into the corridor, other drones still hover at the entrance (where enough structural features exist for their state estimation) and provide passive observation measurements for the first drone. After the first drone flies through the corridor, it hovers at the end, offering passive observation measurements for rest of the drones, so they could also fly through the corridor. Thus, the whole swarm can fly through a smooth corridor without any degeneration. From Fig.~\ref{fig:degeneration2}, it can be seen the proposed method is robust to this type of degenerate scene, the map can recover more structural details than FAST-LIO2. More visual illustrations can be found in our video\footnote{https://youtu.be/MxeoKVXrmEs \label{video}}.

\begin{table}[t]
	\renewcommand\arraystretch{1.2}
	\caption{Localization Accuracy Evaluation}
	\begin{center}
		\scalebox{0.62}{
			\begin{tabular}{p{1.0cm}<{\centering}|p{1.2cm}<{\centering}|p{1.2cm}<{\centering}|p{1.2cm}<{\centering}|p{1.2cm}<{\centering}|p{1.2cm}<{\centering}|p{1.2cm}<{\centering}}
			\toprule
			\multirow{2} *{Data} & \multicolumn{2}{c|}{\textbf{Drone 1}} & \multicolumn{2}{c|}{\textbf{Drone 2}} &  \multicolumn{2}{c}{\textbf{Drone 3}}\\
			\cline{2-7}
			& Swarm-LIO & FAST-LIO2 & Swarm-LIO & FAST-LIO2 & Swarm-LIO & FAST-LIO2 \\
			\cline{1-7}
			01 & 0.0407&\textbf{0.0399} &\textbf{0.0464} &0.0673 &0.0359 &\textbf{0.0347} \\
			\cline{1-7}
			02 &0.0296 &\textbf{0.0289} &\textbf{0.0428} &0.0692 &\textbf{0.0305} &0.0307 \\
			\cline{1-7}
			03 &\textbf{0.0303} &0.0318 &\textbf{0.0442} &0.0836 &\textbf{0.0300} &0.0313 \\
			\cline{1-7}
			04 &0.0355 &\textbf{0.0341} &\textbf{0.0412} &0.0827 &\textbf{0.0298} &0.0310 \\
			\cline{1-7}
			05 &\textbf{0.0242} &0.0255 &\textbf{0.0536} &3.0844 &0.0327 &\textbf{0.0319} \\
			\cline{1-7}
		    \end{tabular}
		}
	\end{center}
	\label{localization_accuracy}
	\begin{tablenotes}
        \footnotesize
        \item
        The position error (RMSE) of FAST-LIO2 and the proposed Swarm-LIO (unit: meter).
    \end{tablenotes}
\end{table}

\subsection{Decentralized Exploration}
\begin{figure}[t]
    \setlength{\abovecaptionskip}{-0.1cm}
	\centering 
	\includegraphics[width=0.45\textwidth]{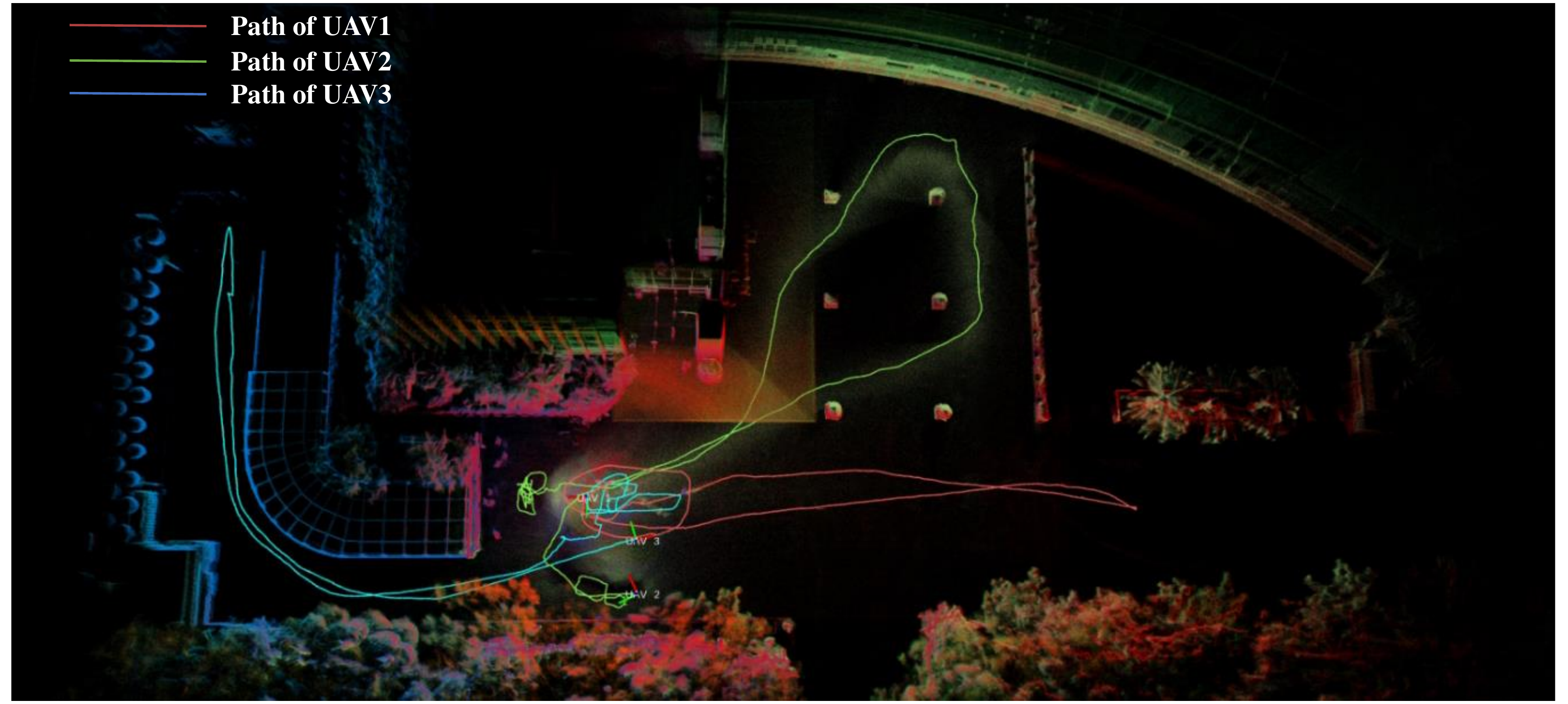}
	\caption{Merged point-cloud map. The red, green, blue points are scanned by drone 1, drone2, drone 3 respectively.}
	\label{fig:exploration}
\end{figure}

To validate that the proposed swarm system is robust to observation lost and can calibrate the global extrinsic transformations, an large-scale decentralized exploration experiment is conducted. 
Three drones are randomly placed in front of a mansion and then take off. After the take-off, they perform some random flights during which each drone detects, tracks, and identifies each other all automatically and without any prior extrinsic information. After that the drones fly in different directions to explore different areas. After flights finish, the point-cloud map of each drone can be merged together offline using the estimated global extrinsic. By transforming point-cloud map of drone 2 and drone 3 to drone 1's global frame, the merged map is shown in Fig.~\ref{fig:exploration}. The consistent merged map shows the capability of accurate global extrinsic estimation without any initial value. Another point should be noted is, since the drones fly in different regions, there is no mutual observation for a long time, \eg, for drone 2 the average distance without mutual observation is \SI{88.58}{\metre}. But still, when the teammate drone returns to drone 2's FoV, the error between the detected teammate position and the predicted position by the teammate tracker (based on the received teammate ego-state, see Sec. \ref{section:teammate_tracking}) is only \SI{0.17}{\metre}, which indicates the high accuracy of the estimated global extrinsic transformation. More visual illustrations are shown in our video.


\section{Conclusion And Future Work}
This paper proposed a fully decentralized and accurate swarm LiDAR-inertial odometry. A novel 3D LiDAR-based drone detection, identification and tracking method was proposed to perform fully autonomous initialization and robust relative state estimation. Mutual observation measurements are tightly-coupled with IMU and point-cloud measurements under ESIKF framework, fulfilling real-time and accurate ego-state estimation, which can keep robust even in degenerate environments for camera or LiDAR. In the future, the drone detection can be upgraded to a auxiliary-free method (\eg, no reflective tape), and the initialization can be more effective. Besides, the scale of the swarm can be larger, more drones will be produced to complete various tasks.

\section{Acknowledgement}
This work is supported in part by the University Grants Committee of Hong Kong General Research Fund under Project 17206421 and in part by DJI under the grant 200009538. The authors gratefully acknowledge Livox Technology for the equipment support and thank Dr. Ximin Lyu of Sun Yat-sen University for the experimental site support during the whole work.

{\small
\bibliographystyle{unsrt}
\bibliography{egbib}
}

\end{document}